\documentclass[runningheads]{llncs}
\usepackage[T1]{fontenc}
\usepackage{graphicx}
\usepackage{xcolor}
\usepackage{hyperref}
\usepackage{cleveref}
\usepackage{url}
\usepackage{tcolorbox}
\tcbuselibrary{listings, skins, breakable}
\begin{document}
\title{Quantifying the Capability Boundary of DeepSeek Models: An Application-Driven Performance Analysis}
%
%

\author{Kaikai Zhao\inst{1,2} \and
Zhaoxiang Liu\inst{*1,2} \and
Xuejiao Lei\inst{1,2} \and
Jiaojiao Zhao\inst{1,2} \and
Zhenhong Long\inst{1,2} \and
Zipeng Wang\inst{1,2} \and
Ning Wang\inst{1,2} \and
Meijuan An\inst{1,2} \and
Qingliang Meng\inst{1,2} \and
Peijun Yang\inst{1,2} \and
Minjie Hua\inst{1,2} \and
Chaoyang Ma\inst{1,2} \and
Wen Liu\inst{1,2} \and
Kai Wang\inst{1,2} \and
Shiguo Lian\inst{*1,2}
}
\authorrunning{Kaikai, Z. et al.}
%
\institute{
Unicom Data Intelligence, China Unicom \and
Data Science \& Artificial Intelligence Research Institute, China Unicom \\
\email
{
\{liansg,liuzx178\}@chinaunicom.cn \\
\inst{*}Corresponding author(s) 
}
}

\maketitle              
\begin{abstract}
DeepSeek-R1, known for its low training cost and exceptional reasoning capabilities, has achieved state-of-the-art performance on various benchmarks. However, detailed evaluations for DeepSeek Series models from the perspective of real-world applications are lacking, making it challenging for users to select the most suitable DeepSeek models for their specific needs. To address this gap, we presents the first comprehensive evaluation of the DeepSeek and its related models (including DeepSeek-V3, DeepSeek-R1, DeepSeek-R1-Distill-Qwen series, DeepSeek-R1-Distill-Llama series, their corresponding 4-bit quantized models, and the reasoning model QwQ-32B) using our enhanced A-Eval benchmark, A-Eval-2.0. Our systematic analysis reveals several key insights: (1) Given identical model architectures and training data, larger parameter models demonstrate superior performance, aligning with the scaling law. However, smaller models may achieve enhanced capabilities when employing optimized training strategies and higher-quality data; (2) Reasoning-enhanced model show significant performance gains in logical reasoning tasks but may underperform in text understanding and generation tasks; (3) As the data difficulty increases, distillation or reasoning enhancements yield higher performance gains for the models. Interestingly, reasoning enhancements can even have a negative impact on simpler problems; (4) Quantization impacts different capabilities unevenly, with significant drop on logical reasoning and minimal impact on text generation. 
Based on these results and findings, we design an model selection handbook enabling users to select the most cost-effective models without efforts. 
It should be noted that, despite our efforts to establish a comprehensive, objective, and authoritative evaluation benchmark, the selection of test samples, characteristics of data distribution, and the setting of evaluation criteria may inevitably introduce certain biases into the evaluation results. We will continuously optimize the evaluation benchmarks and periodically update this paper to provide more comprehensive and accurate evaluation results. Please refer to the latest version of the paper for the most current results and conclusions.
\end{abstract}
\section{Introduction}
In recent years, large language models (LLMs) have revolutionized natural language processing (NLP). Models such as OpenAI’s GPT series \cite{gpt-4o}, Alibaba’s Qwen series \cite{qwen2.5}, MetaAI’s Llama series \cite{dubey2024llama}, and DeepSeek’s DeepSeek series \cite{deepseekai2024deepseekv3technicalreport} have not only advanced NLP technologies but also empowered intelligent solutions for real-world applications. Notably, DeepSeek-R1 \cite{deepseekai2025deepseekr1incentivizingreasoningcapability}, with its extremely low training cost, has achieved state-of-the-art (SOTA) performance comparable to OpenAI’s models \cite{openai-o1mini}. And its exceptional reasoning capabilities have garnered significant attention within the research community. Following the success of DeepSeek-R1, Alibaba introduced QwQ-32B \cite{qwq32b}, a dense reasoning model that also demonstrates competitive performance.

Comprehensive evaluation is essential for understanding model strengths and limitations, as well as guiding optimization and practical deployment. While DeepSeek-R1 \cite{deepseekai2025deepseekr1incentivizingreasoningcapability} has demonstrated SOTA performance on many existing benchmarks such as MMLU \cite{hendrycks2020measuring}, C-Eval \cite{huang2024c}, SimpleQA \cite{simpleqa}, LiveCodeBench \cite{jain2024livecodebench}, Math-500 \cite{math-500}, these benchmarks do not fully capture the nuances of real-world applications. Consequently, three critical challenges arise when deploying DeepSeek models in practical scenarios: First, how should users determine the most suitable model scale? While larger models offer better capabilities, they also come with higher deployment and inference costs. Second, are reasoning-enhanced models always better for all tasks? If not, which tasks are best suited for reasoning-enhanced models, and which tasks can be handled by original instruction-tuned models? Third, is the extent of performance loss observed in the quantized model when applied to various real-world tasks within an acceptable range?

To address the gap, we first create A-Eval-2.0 by modifying some QAs in our previous A-Eval \cite{lian2024best}, an application-driven evaluation benchmark comprising 678 human-curated question-answer (QA) pairs across five major task categories (Text Understanding, Information Extraction, Text Generation, Logical Reasoning, and Task Planning) and 27 subcategories. Then, we evaluate DeepSeek models on A-Eval-2.0. Through this evaluation, we extend existing evaluations and provide actionable insights into how reasoning enhancements influence model performance across various practical tasks. These results and findings help guide users in selecting the most suitable DeepSeek models for their specific needs, enabling cost-effective deployment in real-world applications.

Based on these results and findings, we contribute: (a) The open-sourced A-Eval-2.0 benchmark, including dataset and an evaluation toolkit; (b) The capability boundary quantification results for DeepSeek models; (c) An model selection handbook enabling users to select the most cost-effective models without efforts. Our work provides both methodological contributions to LLM evaluation and practical guidance for real-world applications.

It is important to emphasize that the selection of test samples and the design of evaluation criteria will inevitably introduce certain biases into the evaluation results. To address this, we will continue to optimize this evaluation work to enhance its comprehensiveness and reliability as much as possible.

\section{Evaluation Framework}
\subsection{Evaluated Models}

To assess the impact of reasoning enhancements and quantization in DeepSeek models and assist users in identifying the most cost-effective models for practical applications, we conduct a systematic assessment of 22 models organized into seven distinct groups. These models encompass both Mixture-of-Experts (MoE) and dense models across various series and scales. Each group includes an instruction-tuned model, its corresponding distilled or reasoning-enhanced variant, and the associated quantized model. The instruction-tuned models are from HuggingFace ~\cite{instructmodels-qwen,instructmodels-llama}, the distilled models are from HuggingFace ~\cite{distilledmodels}, and the quantized models are from Ollama ~\cite{quantizedmodels}. \Cref{tab:models} lists the evaluated models and their descriptions. 

\begin{table}[!htbp]
    \centering
    \caption{The Evaluated model list.}
    \begin{tabular}{ll}
    \hline
    \hline
        \textbf{Model} & \textbf{Description} \\ \hline
        Qwen2.5-Math-1.5B-Instruct & Dense math Model, SFT from Qwen2.5-Math-1.5B \\ 
        DeepSeek-R1-Distill-Qwen-1.5B & Dense model distilled from Qwen2.5-Math-1.5B \\
        Distill-Qwen-1.5B-Q4KM & The 4-bit quantized version of distilled model \\ \hline
        Qwen2.5-Math-7B-Instruct & Dense math Model, SFT from Qwen2.5-Math-7B \\ 
        DeepSeek-R1-Distill-Qwen-7B & Dense model Distilled from Qwen2.5-Math-7B \\
        Distill-Qwen-7B-Q4KM & The 4-bit quantized version of distilled model \\ \hline
        Llama-3.1-8B-Instruct & Dense model SFT from Llama-3.1-8B-Base \\ 
        DeepSeek-R1-Distill-Llama-8B & Dense model Distilled from Llama-3.1-8B-Base \\
        Distill-Llama-8B-Q4KM & The 4-bit quantized version of distilled model \\ \hline
        Qwen2.5-14B-Instruct & Dense model SFT from Qwen2.5-14B-Base \\ 
        DeepSeek-R1-Distill-Qwen-14B & Dense model Distilled from Qwen2.5-14B-Base \\
        Distill-Qwen-14B-Q4KM & The 4-bit quantized version of distilled model \\ \hline
        Qwen2.5-32B-Instruct & Dense model SFT from Qwen2.5-32B-Base \\ 
        DeepSeek-R1-Distill-Qwen-32B & Dense model Distilled from Qwen2.5-32B-Base \\
        Distill-Qwen-32B-Q4KM & The 4-bit quantized version of distilled model \\ \hline
        Llama-3.3-70B-Instruct & Dense model SFT from Llama-3.3-70B-Base \\ 
        DeepSeek-R1-Distill-Llama-70B & Dense model Distilled from Llama-3.3-70B-Instruct \\
        Distill-Llama-70B-Q4KM & The 4-bit quantized version of distilled model \\ \hline
        DeepSeek-V3 (671B|37B) & MoE model SFT from DeepSeek-V3-Base \\ 
        DeepSeek-R1 (671B|37B) & MoE model cold-start + RL from DeepSeek-V3-Base \\
        DeepSeek-R1-Q4KM & The 4-bit quantized version of DeepSeek-R1 \\ \hline
        QwQ-32B & Dense model with reasoning ability from Qwen2.5-32B-Base \\ \hline
    \end{tabular}
    \label{tab:models}
\end{table}

\subsection{Dataset}
The A-Eval benchmark \cite{lian2024best} was originally designed to assess the capability boundaries of large language models (LLMs) in practical application scenarios. It consists of 678 manually curated question-answer pairs spanning three difficulty levels, five major categories, and 27 subcategories. While comprehensive for its initial purpose, A-Eval was primarily designed for evaluating conventional instruction-tuned models across different parameter scales.

However, the original benchmark has several limitations in the context of modern LLM development:

1. It does not account for specialized model variants like distilled, quantized, or reasoning-optimized architectures.

2. Its evaluation granularity is insufficient for complex reasoning and planning tasks. 

3. The difficulty progression of data lacks precision in assessing advanced cognitive capabilities.

To address these limitations and better evaluate the enhanced reasoning capacities of DeepSeek models, we introduce A-Eval-2.0. This enhanced version maintains the core structure of A-Eval while incorporating critical improvements in two key aspects:

\textbf{Extended Task Planning Subcategories:}
We add "Scheme Optimization" as a new subcategory under Task Planning category and make 40 new test QA pairs, focusing on evaluating models' ability to execute multi-step plans under constraints and optimize the plans according to dynamic feedback. This sub-task can more fully test the complex planning ability of the reasoning models.

\textbf{Refined Logical Reasoning and Task Planning Categories:}
We refine Logical Reasoning and Task Planning ability into Primary, Intermediate and Advanced levels and make clear definitions:

\textit{Primary Logical Reasoning: } Processes well-defined logical verification or contradiction detection tasks, typically accomplished through simple rules and single-step causal inference.

\textit{Intermediate Logical Reasoning: } Accomplishes multi-factor, multi-step causal reasoning within specific domains by processing conventional constraints and establishing localized logical chains.
 
\textit{Advanced Logical Reasoning: } Addresses domain-specific complex reasoning challenges requiring professional knowledge integration for comprehensive systemic deduction, backtracking, and refinement, ultimately generating professional-grade decisions or solutions.

\textit{Primary Task Planning: } Formulates linearly decomposable deterministic task flows through standardized breakdown of finite steps, executing basic commonsense planning with either explicit parallel branches or no parallelism, without requiring dynamic adjustment.

\textit{Intermediate Task Planning: } Designs multi-constrained comprehensive solutions for single entities by balancing resources, priorities, and personalized needs, while handling common limited variants and optimizing execution paths.

\textit{Advanced Task Planning: } Constructs general-purpose intelligent agents capable of professional-grade, system-level planning in dynamic multi-agent environments, integrating real-time environmental variables to adapt execution strategies.

We use A-Eval-2.0's dataset as the evaluation dataset to analyze the performance of DeepSeek series models in practical applications. On one hand, the evaluation on A-Eval-2.0 enables researchers to assess how the reasoning enhancements in DeepSeek-R1 improve model performance across various practical scenarios. On the other hand, the evaluation results assist users in selecting the most cost-effective models for their specific application needs.

\subsection{Evaluation Process}
We follow the automatic scoring pipeline of original A-Eval, and the automatic scoring process includes:

\textbf{Inference:} Feed each question ${{Q_i}}$ into the evaluated model to generate a prediction ${{P_i}}$.

\textbf{Triplet Preparation:} Construct the triplet $\left( {{Q_i},{A_i},{P_i}} \right)$, where ${{A_i}}$ is the ground-truth answer for ${{Q_i}}$.

\textbf{Scoring:} Combine the prompt and triplet$\left( {{Q_i},{A_i},{P_i}} \right)$, input them into the scoring model, and obtain the scoring output ${{S_i}}$ between 0 and 100. 

To better evaluate the enhanced reasoning capacities of DeepSeek models, we optimize the scoring details in several aspects:

\textbf{Advanced Scoring Model: } We employ DeepSeek-R1 as the scoring model, which demonstrates superior judgment capabilities compared to original instruction-tuned model.

\textbf{Optimized Scoring Prompt: } To better assess the capabilities of instruction-tuned models, distilled models, reasoning models and quantized models, we refine the scoring prompt.

\textbf{Two-Phase Expert Review: } To ensure the reliability of the evaluation results, our evaluation expert team conducts a two-round review of the automatic scoring results:

\textit{Phase 1. Independent Reviewing: } Firstly, $N$ evaluation experts independently assess each automated scoring tuple $\left( {{Q_i},{A_i},{P_i},{S_i}} \right)$ after aligning expert scoring standards. Each expert may output the manual score $E_i^k = {S_i}$ if the automated score ${{S_i}}$ is deemed reasonable or provide the manual score $E_i^k \ne {S_i}$, where $k = \left\{ {1,2,....,N} \right\}$.
For each tuple, we then calculate the consensus ratio:
$C{R_i} = {1 \over N}\sum\limits_{k = 1}^N {{\rm I}\left( {E_i^k = {S_i}} \right)} $
where ${\rm I}$ denotes the indicator function. Finally, ${S_i}$ is considered reliable if $C{R_i} \ge \tau $ (we set $N = 5$ and $\tau  = 0.8$), indicating at least 80\% expert agreement with the automated score ${S_i}$. 

\textit{Phase 2. Consensus Reconciliation: } For tuples failing to meet the consensus threshold ($C{R_i} < \tau $), the evaluation team conducts a focused group discussion. Following discussion, each expert submits a revised score $E_i^{k*}$. The final revised score $S_i^*$ is computed using averaging method that discards the highest and lowest scores.

This two-phase review process ensures both independent assessment and collective wisdom are incorporated into the final evaluation results. The optimized scoring prompt is as follows:

\begin{tcblisting}{
    colback=gray!5,
    colframe=gray!30,
    listing only,
    enhanced,
    boxsep=2pt,
    arc=3pt,
    title=Scoring Prompt Template,
    fonttitle=\small\bfseries,
    listing options={
        basicstyle=\ttfamily\scriptsize,
        breaklines=true
    }
}
You are a rigorous LLM performance evaluator with linguistic expertise and semantic comprehension. Evaluate models' responses against reference answers using this protocol:
1. Analyze question category and requirements
2. Assign score (0-100) based on semantic alignment
--- Evaluation Criteria ---
A. Subjective Tasks (e.g., Task Planning):
   80-100: Complete, error-free response
   60-80: Partially complete with minor omissions
   0-50: Substantially incorrect or irrelevant
B. Objective Tasks (e.g., Classification):
   100: Mathematically equivalent (e.g., -2/3 = -0.666...)
   50: Partially correct (e.g., missing one label)
   0: Completely wrong
--- Special Cases ---
1. Accept linguistically valid alternatives
2. Honor authoritative aliases with justification
3. Allow broader valid terminology
4. Ignore non-semantic formatting differences
5. For Math Reasoning: Recognize mathematically equivalent forms
6. For Task Decomposition: Deduct proportionally for missing elements
7. For Scheme Optimization: Cap at 60 for reflections lacking critique
--- Examples ---
1. [Example case with scoring rationale]
2. [Example case with scoring rationale]
Output format: <score 0-100>
\end{tcblisting}

\section{Results and Discussion}
In this section, we first present the overall average performance of 22 models across all data in A-Eval-2.0 (\Cref{subsection:overall}). Subsequently, we analyze the models' performance from three perspectives: task category (\Cref{subsection:task}), sub-task category (\Cref{subsection:subtask}), and model group (\Cref{subsection:model}). 
Following the A-Eval methodology, we visualize the comparative evaluation results using line charts (\Cref{subsection:linecharts}).
Then, we statistic performance of models on datasets with three difficulty levels—easy, medium, and hard—and discuss models' performance to variations in question difficulty (\Cref{subsection:difficulty}). 
Then, we quantify models' capability boundary through a tier classification table, illustrating performance levels of models on different capabilities (\Cref{subsection:tier}).
Finally, develop a model selection handbook to assist users in selecting the most cost-effective models for their specific requirements without efforts (\Cref{subsection:selection}).

\subsection{Overall Performance}
\label{subsection:overall}
\begin{figure}[hbtp!]
\centerline{\includegraphics[width=1.0\columnwidth]{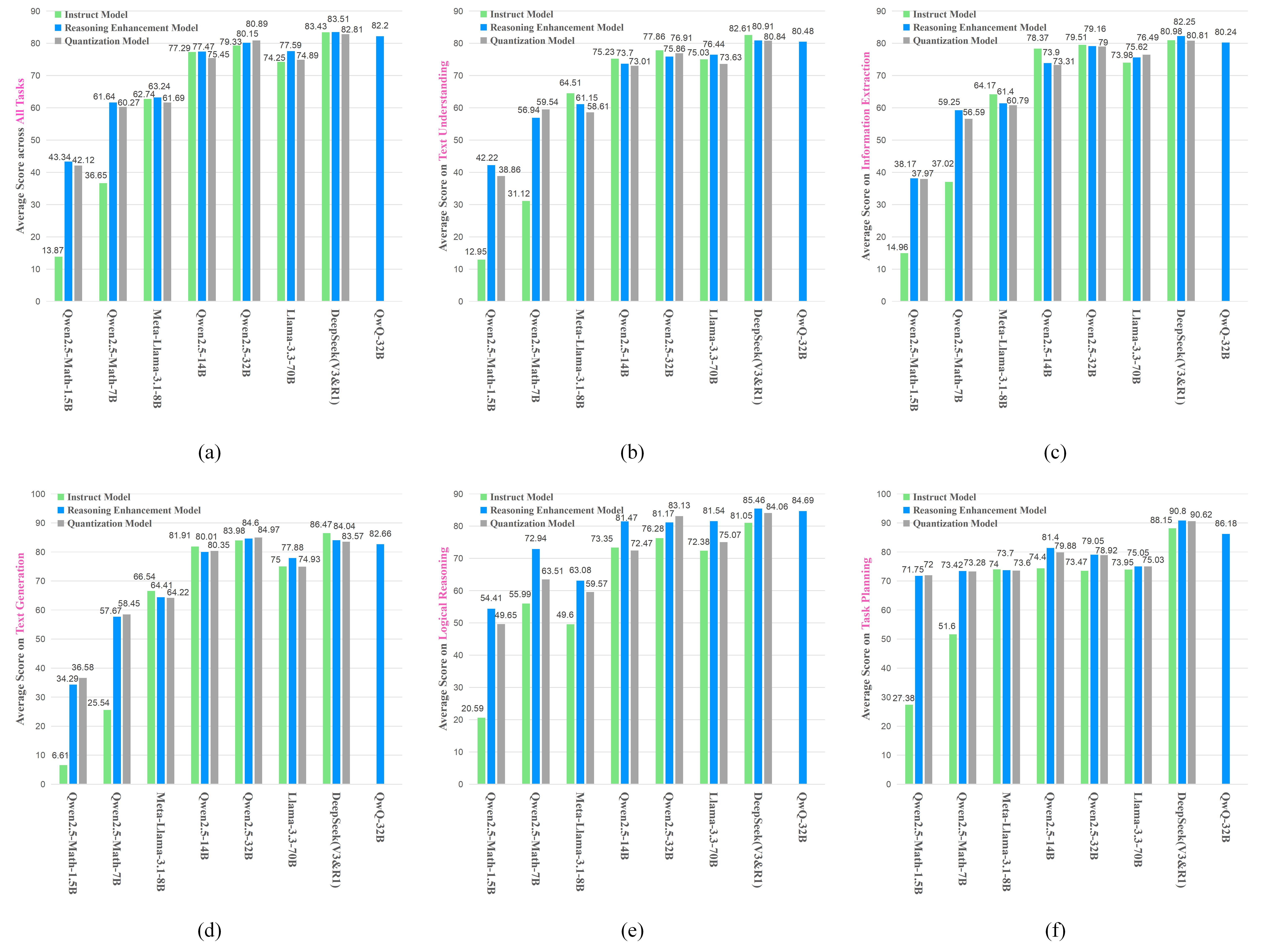}}
\caption{Average score of DeepSeek series models on A-Eval-2.0. (a) The overall average score of models across all data. (b) to (f). The average scores of models on each task. The "Instruct Model" refers to the original Instruction-tuned models without reasoning enhancement, the "Reasoning Enhancement Model" refers to reasoning-enhanced models that have been distilled using DeepSeek-R1's reasoning data or the native reasoning models, and the "Quantization Model" refers to 4-bit quantized version of the reasoning enhancement models.}
\label{fig:scores_eachmodel_task}
\end{figure}

\Cref{fig:scores_eachmodel_task} (a) presents the overall average scores of each model group across all data.

\textbf{Findings Consistent with Common Sense: }

\textcolor{green}{\textbf{(1)}} Undoubtedly, DeepSeek-V3 and DeepSeek-R1 demonstrate the best overall performance among all evaluated models. The 4-bit quantized version of DeepSeek-R1 demonstrates only a 0.84\% performance degradation compared to the full-precision version, while still outperforming both Qwen and Llama series models.

\textcolor{green}{\textbf{(2)}} Overall, reasoning-enhanced models outperform their original instruction-tuned counterparts. After 4-bit quantization, the model's performance decreases by an average of 1.93\%.

\textcolor{green}{\textbf{(3)}} Overall, given identical model architectures and training data, larger parameter models demonstrate superior performance, aligning with the scaling law \cite{gao2023scaling,kaplan2020scaling}.

\textcolor{green}{\textbf{(4)}} Distillation brings the most significant improvements to Qwen2.5-Math-1.5B and Qwen2.5-Math-7B, with score increases of 212.47\% and 68.19\%, respectively. This is because they are math-focused and perform poorly on general tasks, but the distilled reasoning data from DeepSeek-R1 significantly enhances their general capabilities.

\textbf{Findings Contrary to Common Sense:}

\textcolor{red}{\textbf{(5)}} Both before and after distillation, Qwen2.5-32B consistently outperforms the larger Llama-3.3-70B model. The counterintuitive result is likely due to the predominance of Chinese data in our evaluation dataset, and this phenomenon also appears later in the discussion of performance comparison by task (\Cref{subsection:task}) and by subtask (\Cref{subsection:subtask}).

\textcolor{red}{\textbf{(6)}} Smaller models can achieve enhanced capabilities when employing optimized training strategies and higher-quality data. For instance, the reasoning model QwQ-32B outperforms the original instruction-tuned model Qwen2.5-32B-Instruct by 3.62\% in overall performance, surpasses the distilled model DeepSeek-R1-Distill-Qwen-32B by 2.56\%, and trails the DeepSeek-R1 by only 1.57\%.

\subsection{Performance Comparison By Task}
\label{subsection:task}
\Cref{fig:scores_eachmodel_task}(b) to ~\Cref{fig:scores_eachmodel_task} (f) compare the scores of the models on the five major task categories.

\textbf{Findings:}

\textcolor{green}{\textbf{(7)}} The two mathematical models achieve comparable performance in Logical Reasoning and Task Planning but under-perform in other tasks.

\textcolor{green}{\textbf{(8)}} Compared to instruction-tuned model DeepSeek-V3, reasoning model DeepSeek-R1 achieves significant gains of 5.4\% and 3.0\% in logical reasoning and task planning tasks, respectively. However, it demonstrates performance decreases of 2.1\% and 1.6\% in text understanding and generation tasks. 

\textcolor{green}{\textbf{(9)}} When distilled using DeepSeek-R1's data, Qwen and Llama model families show overall performance improvements of 1.63\% and 2.65\% respectively. Particularly in logical reasoning tasks, the distilled models achieve substantial gains of 9.50\% and 19.92\% respectively. However, Performance degradation occurs after distillation in: Text Understanding (Llama-3.1-8B, Qwen2.5-14B, Qwen2.5-32B, and DeepSeek-V3), Information Extraction (Llama-3.1-8B, Qwen2.5-14B), and Text Generation (Llama-3.1-8B, Qwen2.5-14B, DeepSeek-V3), and Task Planning (Llama-3.3-70B).

\textcolor{green}{\textbf{(10)}} Quantization affects different task types variably: Minimal impact on text generation and task planning (average performance drop: 0.32\%); Moderate impact on text comprehension and information extraction (average drops: 1.58\% and 1.09\%); Significant impact on logical reasoning (average drop: 6.49\%)

\begin{figure}[!htpb]
\centerline{\includegraphics[width=0.9\columnwidth]{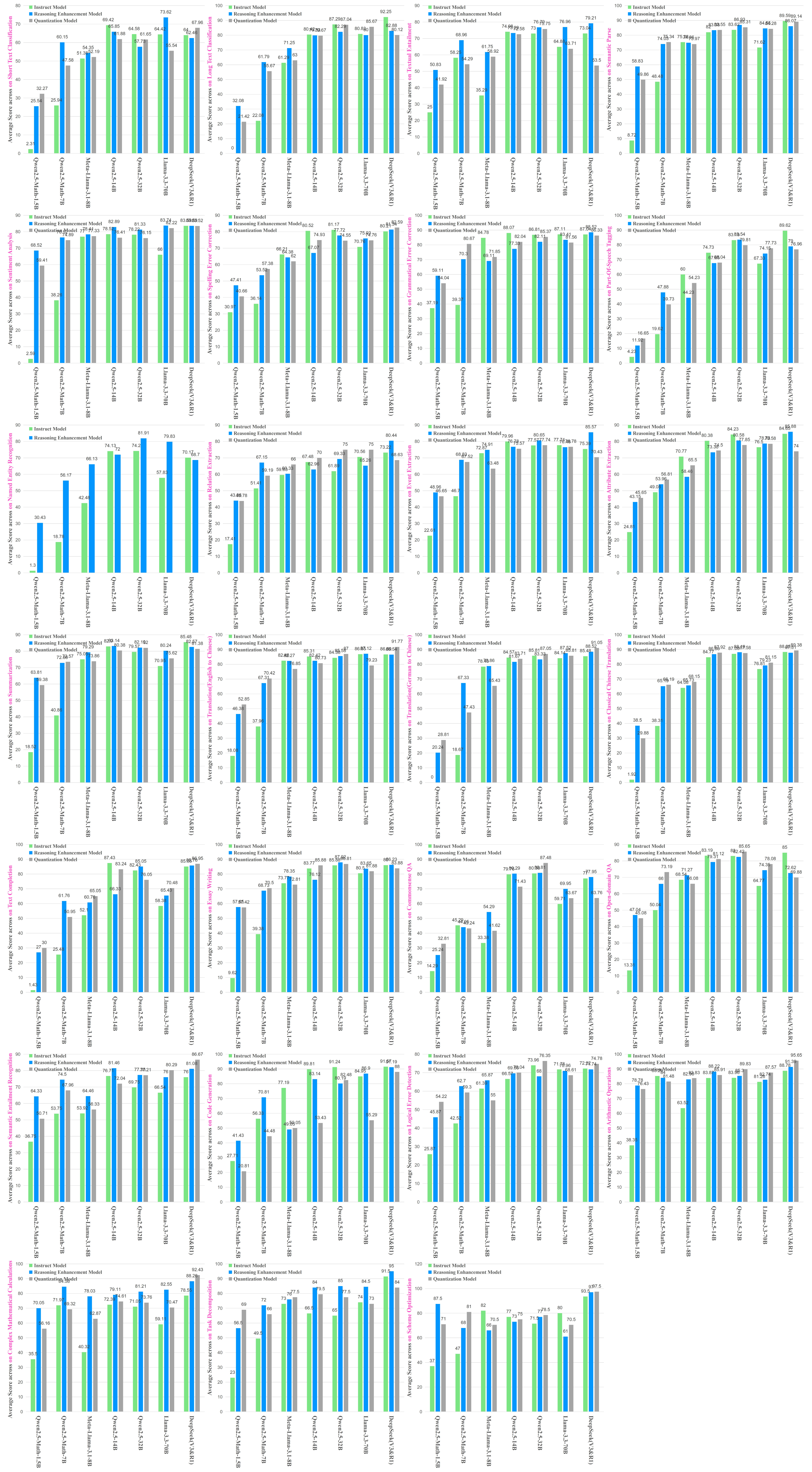}}
\caption{Average scores of the DeepSeek series models on the 27 subcategories.}
\label{fig:scores_eachmodel_subtask}
\end{figure}

\subsection{Performance Comparison By Subtask}
\label{subsection:subtask}
In more detail, \Cref{fig:scores_eachmodel_subtask} provides the scores of the models on the 27 subcategories. 

After analyzing these results, here are some interesting findings:

\textcolor{green}{\textbf{(11)}} DeepSeek models achieve dominant performance in 21 out of 27 subtasks. However, QwQ-32B demonstrates superior capability in the remaining six subtasks: Named Entity Recognition, Event Extraction, Common Sense QA, Open-domain QA, Semantic Entailment Recognition, and Code Generation.

\textcolor{green}{\textbf{(12)}} Compared to other tasks, distillation brings the highest gains in Complex Mathematical Computation subtask, with an average improvement of 62.86\%.

\subsection{Performance Comparison By Model}
\label{subsection:model}
\begin{figure}[!htpb]
\centerline{\includegraphics[width=1\columnwidth]{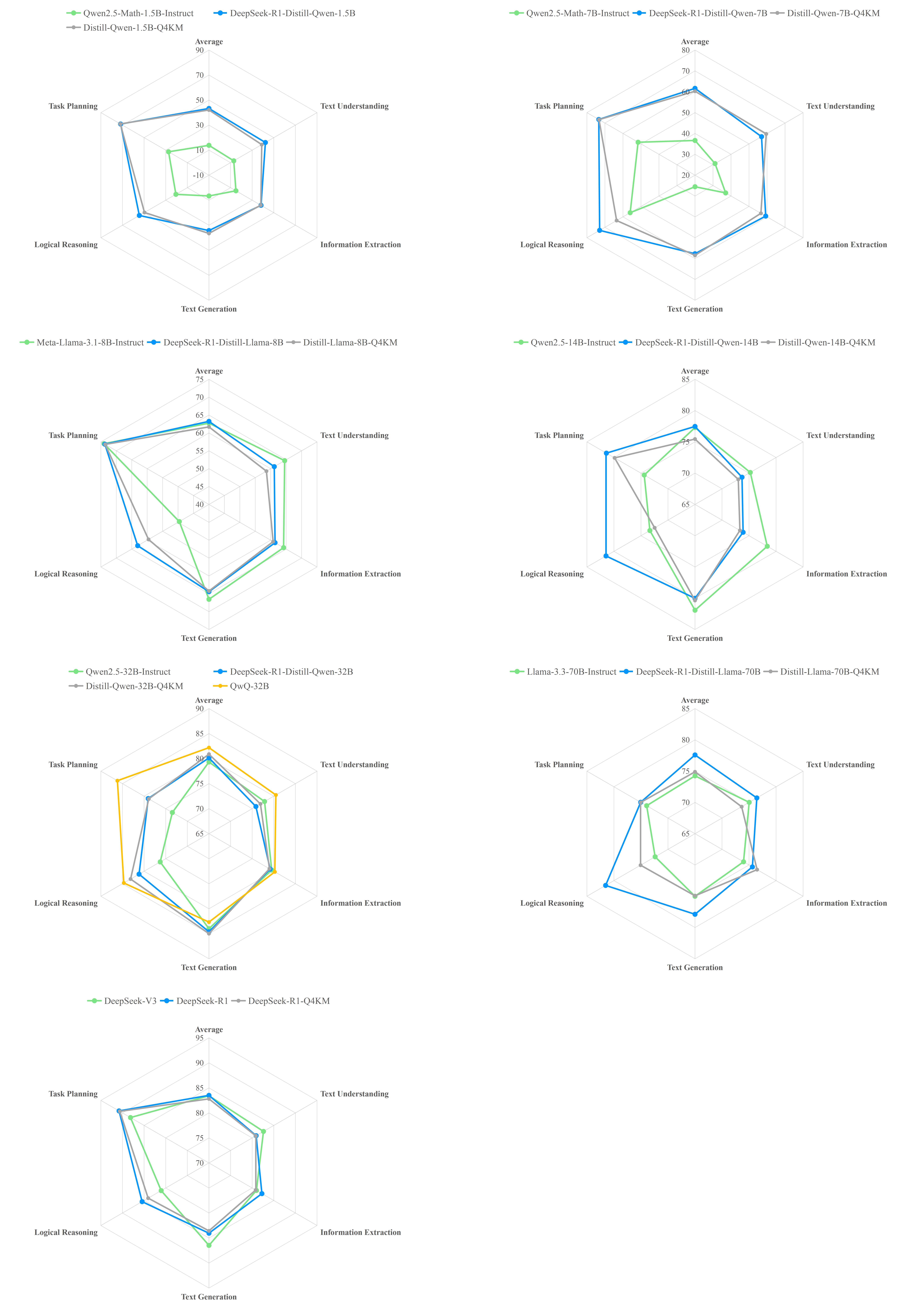}}
\caption{Performance of each model group on five major tasks.}
\label{fig:scores_bymodel}
\end{figure}

To more clearly compare the performance of each group of models before and after reasoning enhancement on the five major tasks, we present the evaluation scores by seven model groups in \Cref{fig:scores_bymodel}.

\subsection{Line Chart Results}
\label{subsection:linecharts}
we present the evaluation results using line charts, following the A-Eval methodology. These visualizations provide a clear and intuitive comparison of model performance across different tasks and subcategories.

\begin{figure}[!htpb]
\centerline{\includegraphics[width=1\columnwidth]{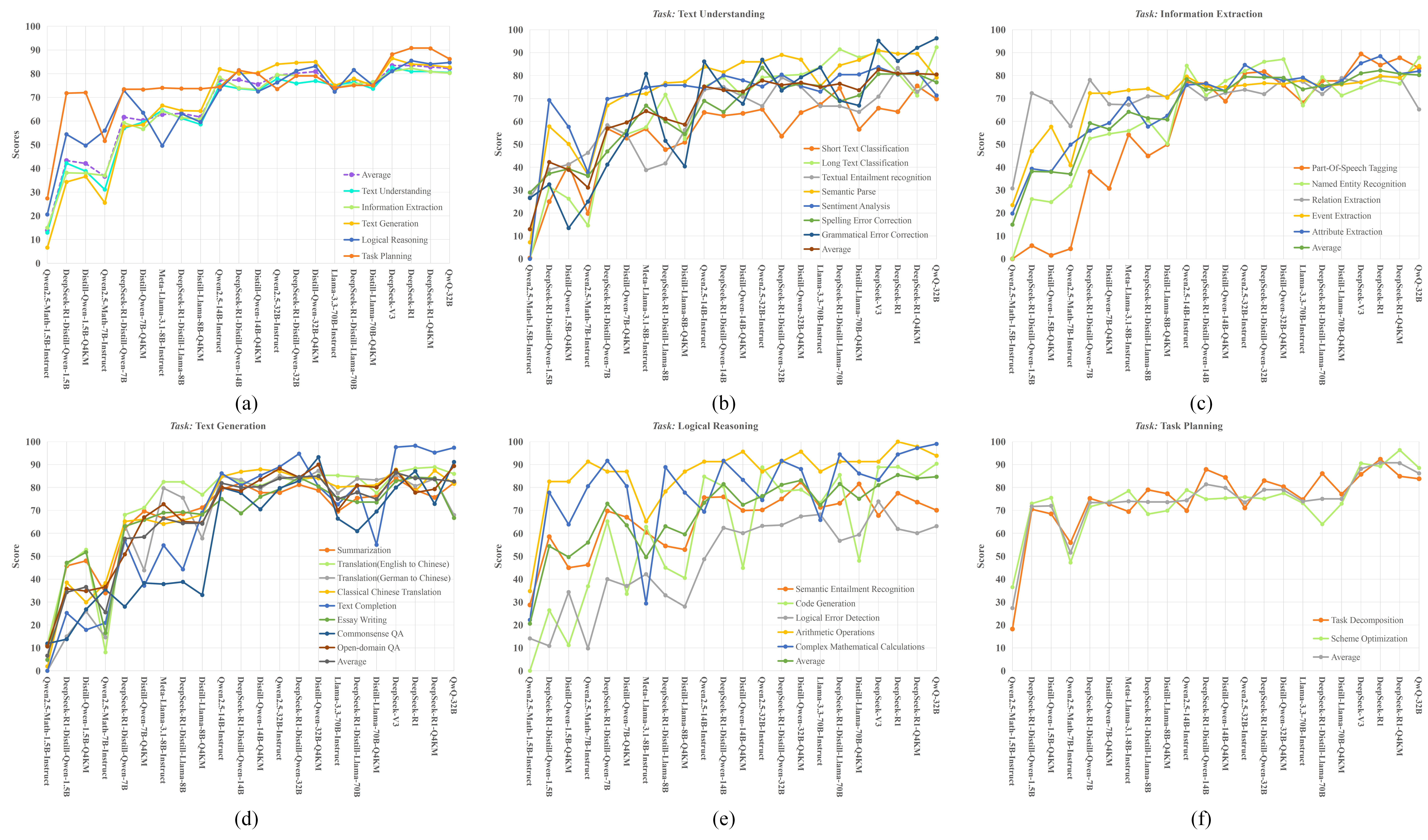}}
\caption{Line charts for evaluation performance. (a) Line chart on five major tasks. (b) - (f) Line charts on five major tasks and corresponding subtasks}
\label{fig:linechart}
\end{figure}

\Cref{fig:linechart} (a) illustrates the scores of the 22 evaluated models on each major task, as well as their average scores across all tasks. This comprehensive overview allows users to quickly identify models that excel in specific major task types or demonstrate strong overall performance.

In \Cref{fig:linechart} (b) to \Cref{fig:linechart} (f), we further break down the performance of the 22 models by showing their average scores within each of the five major task categories and their corresponding subcategories. These detailed visualizations enable users to assess model capabilities at a granular level, ensuring a precise match between model strengths and task requirements.






\subsection{Performance Comparison By Difficulty Level}
\label{subsection:difficulty}
In A-Eval and A-Eval-2.0, each QA pair is assigned a difficulty level label: Easy, Medium, or Hard. It is evident that models tend to achieve lower scores as the difficulty of the questions increases. However, there has been no quantitative analysis of how DeepSeek models perform on questions of varying difficulty levels. To address this gap, we separately analyze the evaluation results for each difficulty level and present our findings.

\begin{figure}[!htpb]
\centerline{\includegraphics[width=1.0\columnwidth]{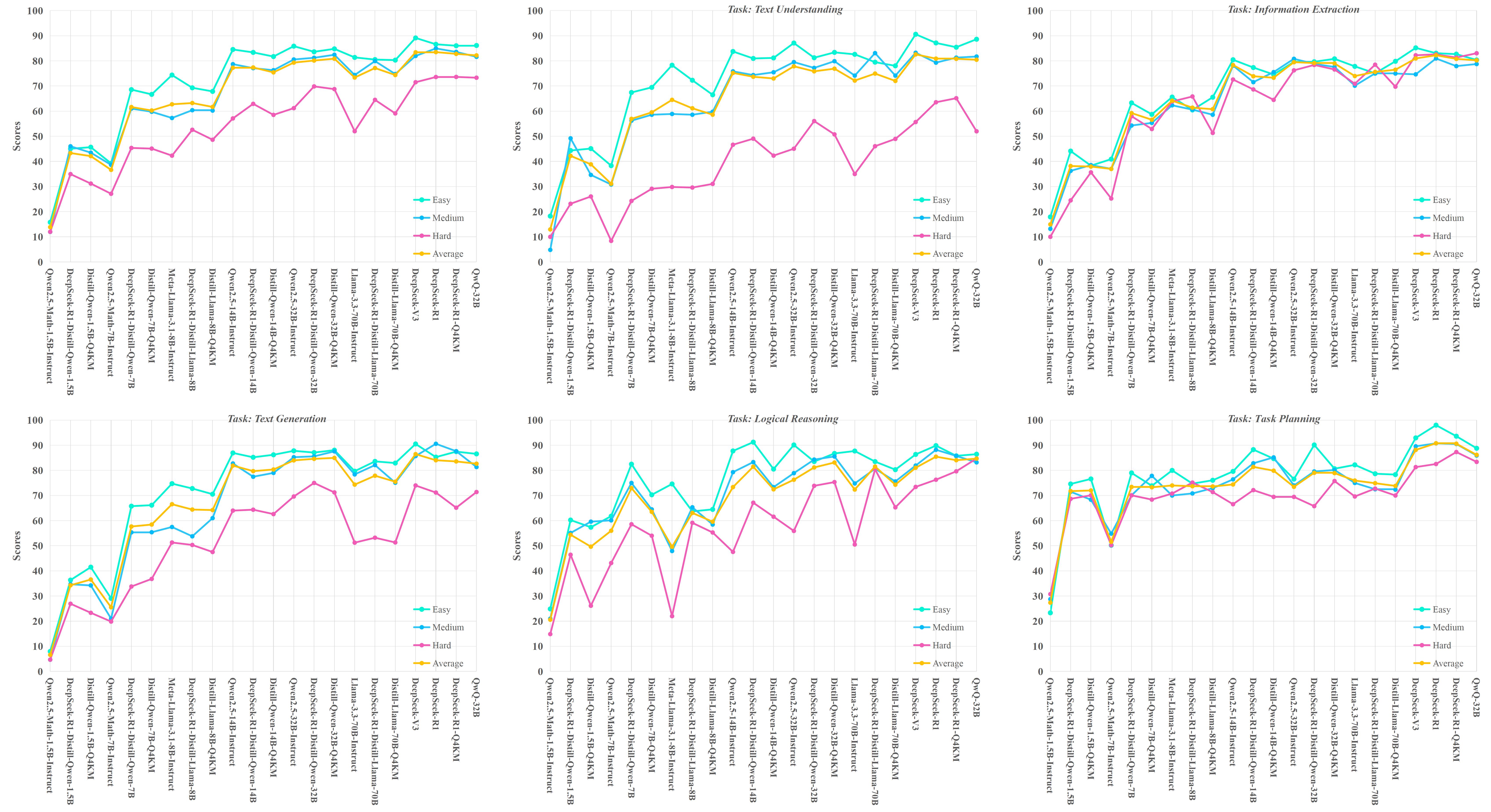}}
\caption{Average performance of 22 models on Easy, Medium, and Hard data.}
\label{fig:scores_difficulty_overallandtask}
\end{figure}

\Cref{fig:scores_difficulty_overallandtask} (a) illustrates the average scores of 22 models across task categories for Easy, Medium, and Hard data. Meanwhile, \Cref{fig:scores_difficulty_overallandtask} (b) - (f) depict the scores of the 22 models for each task category across the three difficulty levels.

\begin{table}[!ht]
    \centering
    \caption{Performance tier classification of DeepSeek models across major task categories. Models are categorized into tiers (A+: >85, A: 80-85, B: 75-80, C: 60-75, D: <60) based on their scores in Text Understanding (TU), Information Extraction (IE), Text Generation (TG), Logical Reasoning (LR), and Task Planning (TP).}
    \begin{tabular}{|c|c|c|c|c|c|c|}
    \hline
        \textbf{Model} & \textbf{Model Scale} & \textbf{  TU  } & \textbf{  IE  } & \textbf{  TG  } & \textbf{  LR  } & \textbf{  TP  } \\ \hline
        Qwen2.5-Math-1.5B-Instruct & 1.5B & D & D & D & D & D \\
        DeepSeek-R1-Distill-Qwen-1.5B & 1.5B & D & D & D & D & C \\
        Distill-Qwen-1.5B-Q4KM & 1.5B & D & D & D & D & C \\ \hline
        Qwen2.5-Math-7B-Instruct & 7B & D & D & D & D & D \\
        DeepSeek-R1-Distill-Qwen-7B & 7B & D & D & D & C & C \\
        Distill-Qwen-7B-Q4KM & 7B & D & D & D & C & C \\ \hline
        Llama-3.1-8B-Instruct & 8B & C & C & C & D & C \\ 
        DeepSeek-R1-Distill-Llama-8B & 8B & C & C & C & C & C \\
        Distill-Llama-8B-Q4KM & 8B & D & C & C & D & C \\ \hline
        Qwen2.5-14B-Instruct & 14B & B & B & A & C & C \\ 
        DeepSeek-R1-Distill-Qwen-14B & 14B & C & C & A & A & B \\
        Distill-Qwen-14B-Q4KM & 14B & C & C & A & C & B \\ \hline
        Qwen2.5-32B-Instruct & 32B & B & B & A & B & C \\
        DeepSeek-R1-Distill-Qwen-32B & 32B & B & B & A & A & B \\
        Distill-Qwen-32B-Q4KM & 32B & B & B & A & A & B \\ \hline
        Llama-3.3-70B-Instruct & 70B & A & B & A & B & B \\ 
        DeepSeek-R1-Distill-Llama-70B & 70B & A & A & A & A & A \\
        Distill-Llama-70B-Q4KM & 70B & B & A & B & A & A \\ \hline
        DeepSeek-V3 & 671B|37B & A & A & A+ & A & A+ \\ 
        DeepSeek-R1 & 671B|37B & A & A & A & A+ & A+ \\
        DeepSeek-R1-Q4KM & 671B|37B & A & A & A & A & A+ \\ \hline
        QwQ-32B & 32B & A & A & A & A & A \\ \hline
    \end{tabular}
    \label{tab:selection}
\end{table}



\begin{figure}[!htpb]
\centerline{\includegraphics[width=1\columnwidth]{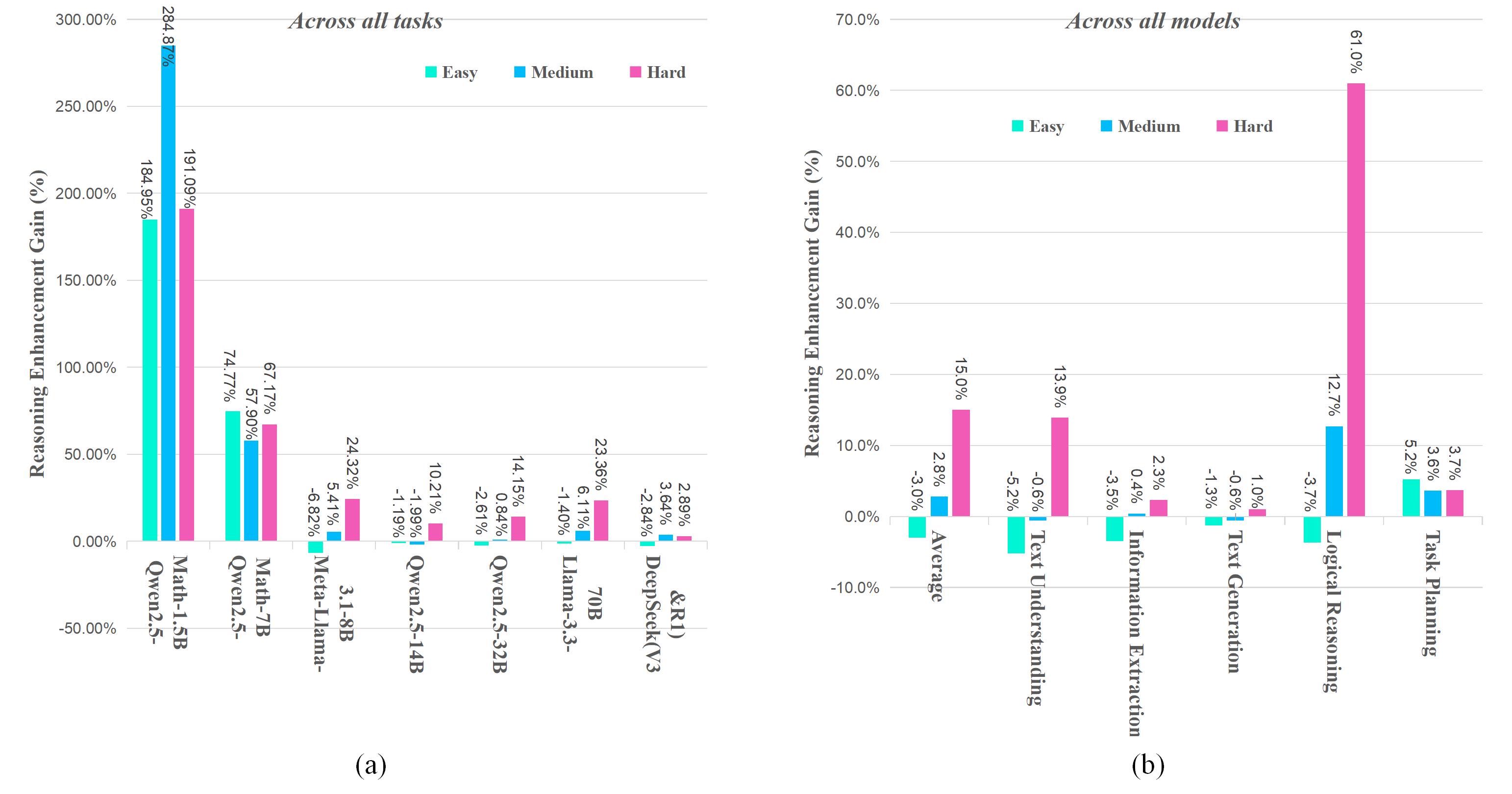}}
\caption{Performance gain of reasoning enhancement on Easy, Medium, and Hard data. (a) Performance gain of each model group across all tasks. (b) Performance gain of each task category across all models (excluding math models).}
\label{fig:enhancement_difficulty}
\end{figure}

In \Cref{fig:enhancement_difficulty}, we analyze the percentage score gains brought by reasoning enhancements on Easy, Medium, and Hard data. An interesting observation emerges:

\textcolor{green}{\textbf{(13)}} As the data difficulty increases, distillation or reasoning enhancements yield higher performance gains for the models. Interestingly, reasoning enhancements can even have a negative impact on simpler problems.

\subsection{Model Performance Tier Classifications}
\label{subsection:tier}
We classify model performance into five tiers (A+, A, B, C, D) based on their scores in the five major task categories, resulting in the capability boundary quantification table (\Cref{tab:selection}).

As shown in \Cref{tab:selection}, models are assigned a tier for each task category, with A+ representing the highest performance (scores > 85) and D indicating the lowest (scores < 65). This tiered classification provides an intuitive and accessible way for users to quickly identify models' performance for specific abilities.

\subsection{Model Selection Notebook for Users}
\label{subsection:selection}
To assist users in selecting the most suitable model based on their specific application requirements, we develop a model selection notebook consisting of four components.

\begin{figure}[!htpb]
\centerline{\includegraphics[width=1\columnwidth]{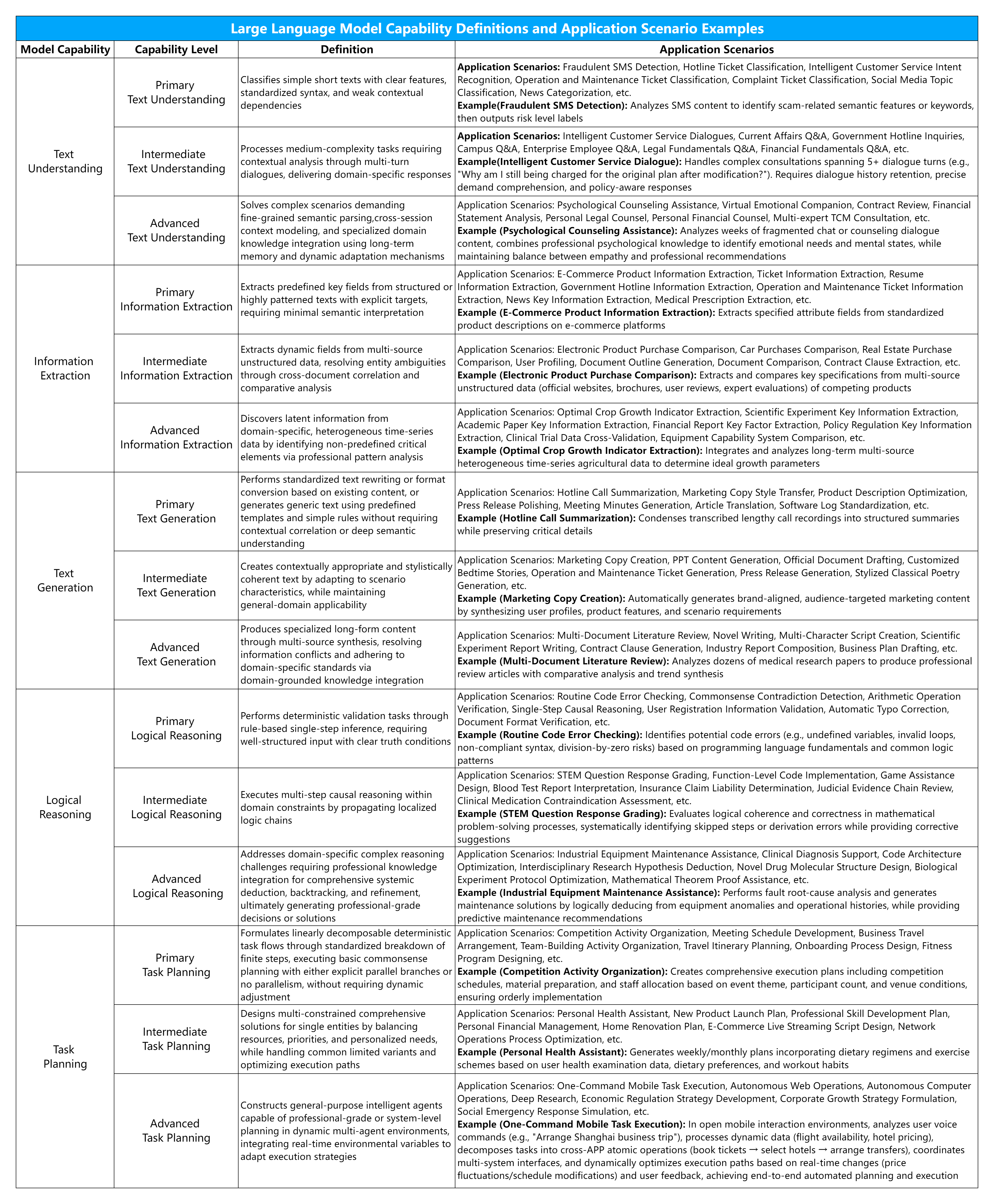}}
\caption{LLM capability definitions and application scenarios, including 15 well-defined model capabilities along with their corresponding 105 application scenarios.}
\label{fig:definitions}
\end{figure}

\textbf{1. LLM Capability Definitions and Application Scenarios.} As shown in \Cref{fig:definitions}, we systematically categorize a substantial number of real-world application scenarios and analyze the specific model capabilities required for each. 

\begin{figure}[!htpb]
\centerline{\includegraphics[width=1\columnwidth]{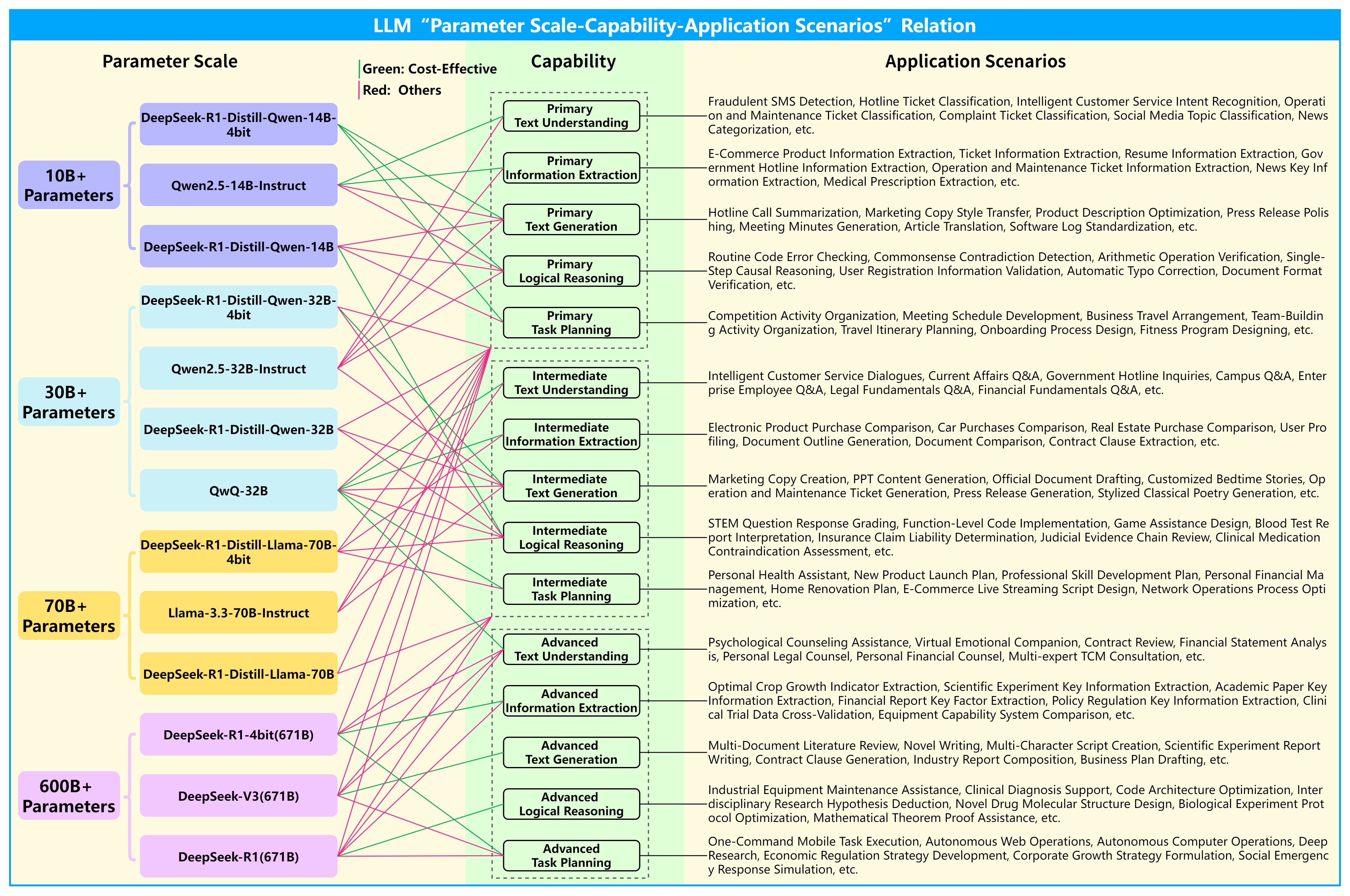}}
\caption{LLM “Parameter Scale-Capability-Application” relation. The figure illustrates the correlations among models, their capabilities, and practical applications. For each capability, the most cost-effective model is indicated by green lines, while other models with higher computational costs are shown in red.}
\label{fig:relation}
\end{figure}

\textbf{2. LLM “Parameter Scale-Capability-Application Scenarios” Relation.} As shown in \Cref{fig:relation}, we analyze the relationship among parameter scale, capability, and application based on the evaluation results and the model performance tier classifications in \Cref{tab:selection}. 

\begin{figure}[!htpb]
\centerline{\includegraphics[width=1\columnwidth]{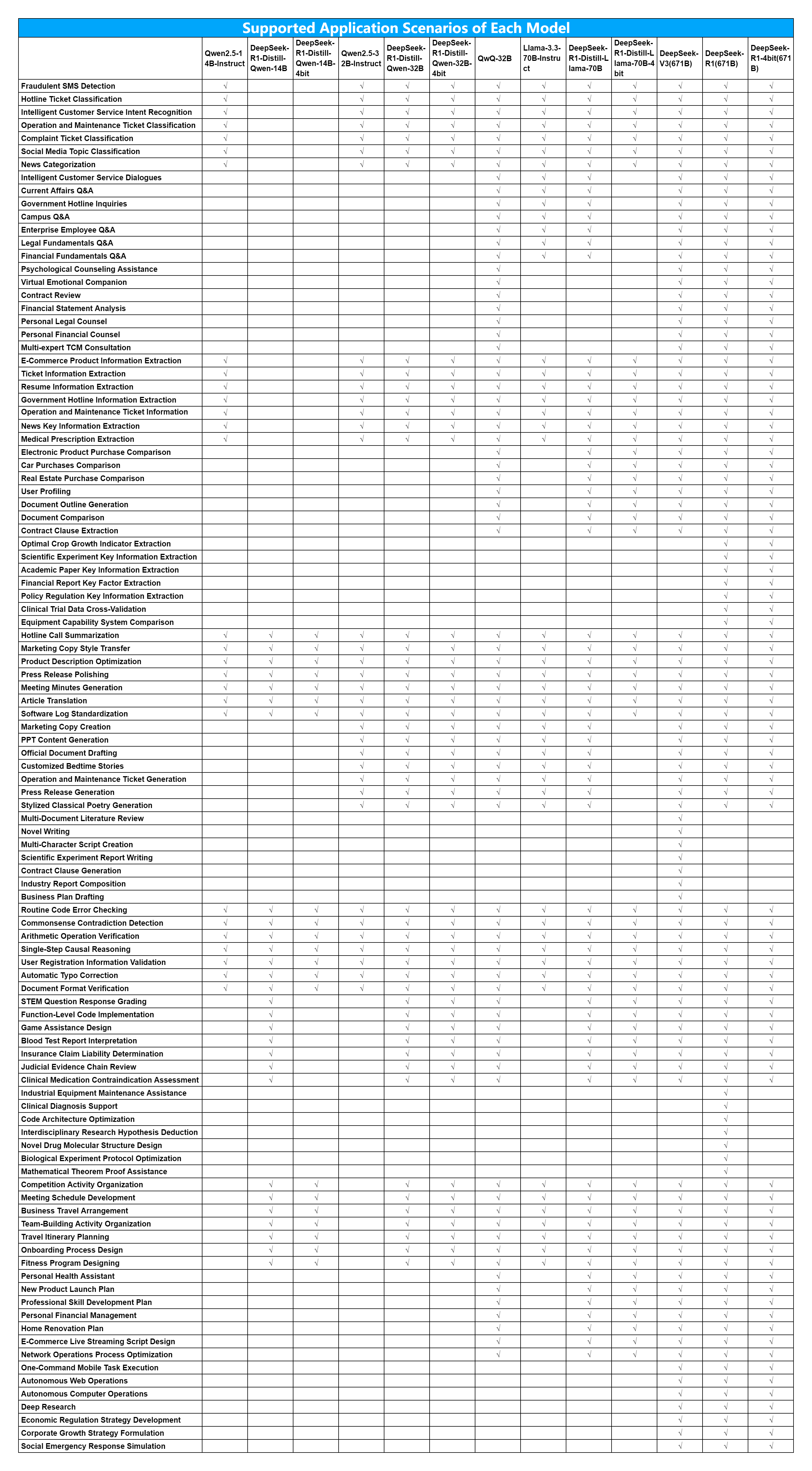}}
\caption{Supported application scenarios of each model.}
\label{fig:supported}
\end{figure}

\textbf{3. Supported Application Scenarios for Each Model.} As shown in \Cref{fig:supported}, we clearly demonstrate the specific application scenarios that each model can successfully accomplish.

\begin{figure}[!htpb]
\centerline{\includegraphics[width=1\columnwidth]{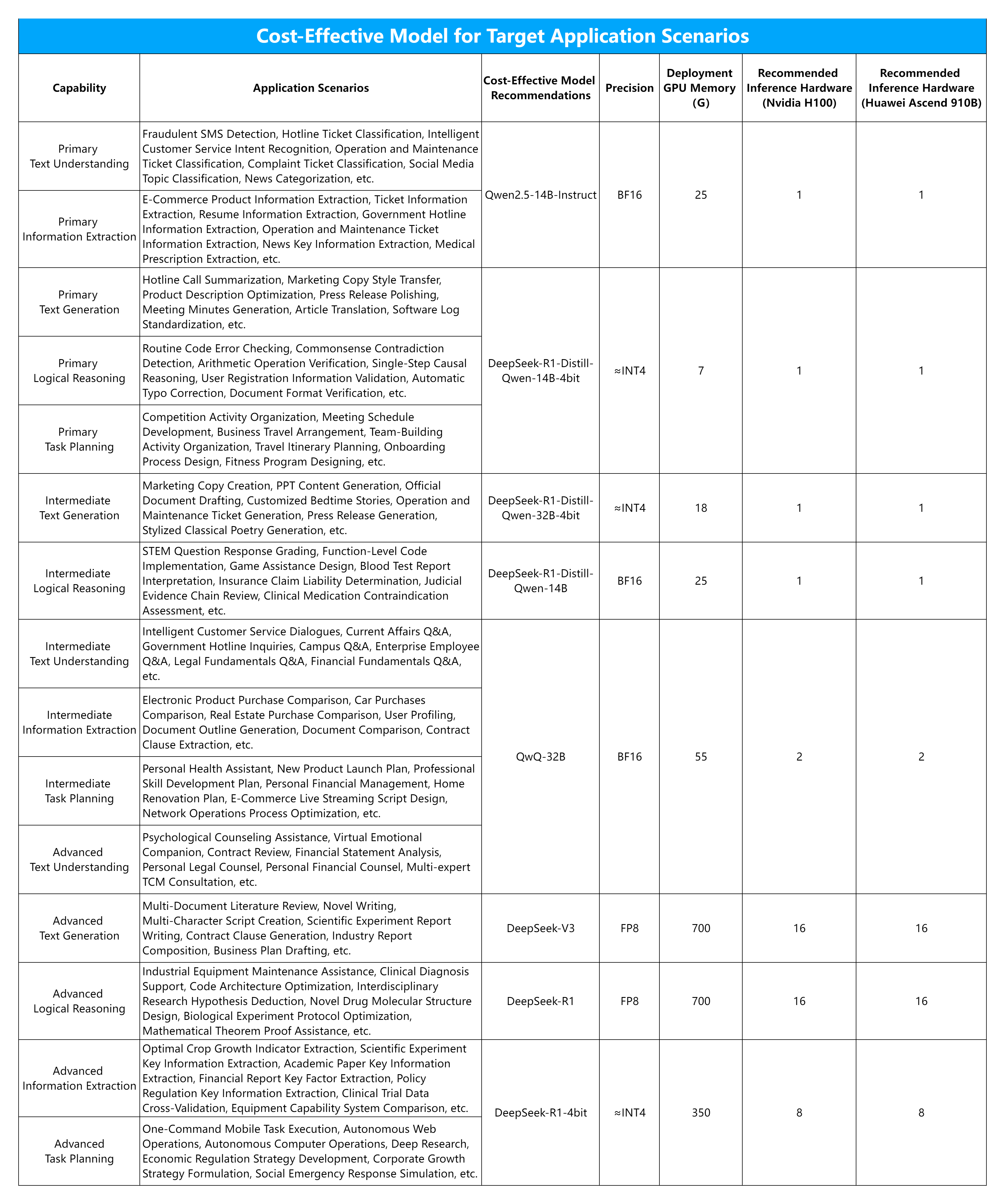}}
\caption{Cost-Effective model recommendations for target application scenarios.}
\label{fig:costeffective}
\end{figure}

\textbf{4. Cost-Effective Model Recommendations for Target Application Scenarios.} As illustrated in \Cref{fig:costeffective}, to simplify the model selection process for users without extensive knowledge of LLMs, we provide cost-effective model recommendations for each application scenario, along with detailed hardware deployment requirements.

Using our model selection notebook, users can efficiently identify the most cost-effective models, ensuring optimal performance and resource efficiency in real-world applications.

It should be noted that, despite our efforts to establish a comprehensive, objective, and authoritative evaluation benchmark, the selection of test samples, characteristics of data distribution, and the setting of evaluation criteria may inevitably introduce certain biases into the evaluation results. We will continuously optimize the evaluation benchmarks and periodically update this paper to provide more comprehensive and accurate evaluation results. Please refer to the latest version of the paper for the most current results and conclusions.
\newpage
\section{Conclusion}
To better understand how DeepSeek models perform in real-world applications, we comprehensively evaluate DeepSeek models, their distilled variants, 4-bit quantized models, and QwQ-32B on the modified A-Eval-2.0 benchmark. Our analysis reveals that reasoning-enhanced models, while generally powerful, are not universally superior across all tasks. Additional, we quantify the capability boundary of DeepSeek models through performance tier classifications. Finally, we develop the model selection handbook, providing actionable insights to help users select and deploy the most cost-effective models based on their specific application requirements. In the future, we will continue to advance this work by optimizing evaluation benchmarks and promptly updating evaluation results to the community.
%
\bibliographystyle{splncs04}
\bibliography{mybibliography}

\end{document}